\documentclass[preprint,review,12pt]{elsarticle}

\usepackage{graphicx}
\usepackage{amsmath}
\usepackage{amssymb}
\usepackage{subfigure}
\usepackage{multicol,multirow,booktabs}
\usepackage{caption}
\usepackage{algorithm}
\usepackage{verbatim}
\journal{Pattern Recognition}

\begin{document}

\begin{frontmatter}



\title{K-means clustering for efficient and robust registration of multi-view point sets}


\author[add1]{Zutao Jiang}
\author[add1]{Jihua Zhu\corref{mycorresponding author}}
\cortext[mycorresponding author]{Corresponding author: Jihua Zhu}
\ead{zhujh@xjtu.edu.cn}
\author[add2]{Georgios D. Evangelidis}
\author[add3]{Changqing Zhang}
\author[add1]{Shanmin Pang}
\author[add1]{Yaochen Li}

\address[add1]{School of Software Engineering, Xi'an Jiaotong Universiy, P. R. China}
\address[add2]{DAQRI, Vienna, Austria}
\address[add3]{School of Computer Science and Technology, Tianjin Universiy, P. R. China}

\begin{abstract}
Generally, there are three main factors that determine the practical usability of registration, i.e., accuracy, robustness, and efficiency. In real-time applications, efficiency and robustness are more important. To promote these two abilities, we cast the multi-view registration into a clustering task. All the centroids are uniformly sampled from the initially aligned point sets involved in the multi-view registration, which makes it rather efficient and effective for the clustering. Then, each point is assigned to a single cluster and each cluster centroid is updated accordingly. Subsequently, the shape comprised by all cluster centroids is used to sequentially estimate the rigid transformation for each point set. For accuracy and stability, clustering and transformation estimation are alternately and iteratively applied to all point sets. We tested our proposed approach on several benchmark datasets and compared it with state-of-the-art approaches. Experimental results validate its efficiency and robustness for the registration of multi-view point sets.
\end{abstract}


\end{frontmatter}


\section{Introduction}\label{sec:introduction}

As a fundamental issue in many areas, the problem of point set registration has attracted immense attention in computer vision \cite{shiratori2015efficient,yang2016go,Liu2018Global}, computer graphics \cite{dai2017bundlefusion,aiger20084}, and robotics \cite{yu2015semantic,ma2016merging}. The aim of point set registration is to estimate the optimal transformation between different point sets. According to the involved number of point sets, this problem can be roughly divided into the pair-wise registration and the multi-view registration, with the former being extensively addressed.

Commonly, the problem of pair-wise registration is solved by the iterative closest point (ICP) algorithm \cite{besl1992method} or its variants \cite{chetverikov2005robust,phillips2007outlier,langis2001parallel}. To achieve registration, most of these approaches alternately build hard correspondences and estimate the transformation. Such approaches are efficient but may be not accurate enough. Furthermore, some registration approaches \cite{granger2002multi,jian2011robust,myronenko2010point,tsin2004correlation} replace the hard assignment with the soft assignment so as to obtain more accurate results. Since the soft assignment should be built from each point to all points in the opposite point set, these approaches are time-consuming. Besides, most of ICP variants are locally convergent. To obtain the desired global minimum, particle filter \cite{sandhu2010point,kolesov2016stochastic} and genetic algorithm \cite{lomonosov2006pre,zhu2014robust} can be combined to estimate the optimal transformation. 3D features can be also extracted from the point sets and they can be matched to provide initial transformation for the pair-wise registration\cite{rusu2009fast,lei2017fast,lou2014image}.

The development of scanning equipment makes the 3D reconstruction of an object possible. Due to the occlusion, however, the object cannot be entirely scanned from a single viewpoint. Therefore, scanners should acquire point sets from different viewpoints so as to cover the entire object surface. These point sets can then be transformed into one common reference frame for the 3D model reconstruction\cite{ren2017reconstruction}.

Unlike pair-wise registration, multi-view registration is more difficult and has comparatively attracted less attention. Usually, the multi-view registration can be divided into two stages: the coarse registration and fine registration. For the coarse registration, some approaches apply the pair-wise registration method on all or some of point set pairs, then search the minimum set of pair-wise registration results to estimate the initial rigid transformations for the multi-view registration \cite{huber2003fully,mian2006three,zhu2016automatic}. Since there are a lot of point set pairs involved in the multi-view registration, the coarse registration approaches are always time-consuming. To accelerate this stage, 3D features can be extracted and matched to provide good initial parameters for the pair-wise registration \cite{fantoni2012accurate, guo2014accurate, guo2015integrated}.

The fine registration has attracted considerably more attention. The common strategy is to sequentially align and merge two point sets until all point sets are merged into one model \cite{chen1992object}. Although this approach may be efficient, it generally suffers from the error accumulation. To address this issue, Bergevin \emph{et al}. \cite{bergevin1996towards} suggested to establish the correspondence between one point set and any other point set so as to estimate the rigid transformation of this point set. As the establishment of correspondence is time-consuming, the efficiency of this approach is very low. Therefore, Pulli \cite{pulli1999multiview} proposed to align point sets with each other and utilize the pairwise alignments as constraints to evenly diffuse the pairwise registration errors in the multi-view registration step. Inspired by these approaches, the pair-wise registration algorithm can be sequentially utilized to align one point set to the coarse model constructed by all point sets \cite{zhu2014surface}. The outcome can be used to update the coarse model, which can be further used to align each point set.
Due to too many points in the reconstructed model, however, the complexity of these approaches is relatively high. The multi-view registration can be also viewed as graph optimization problem \cite{sharp2004multiview,shih2008efficient,torsello2011multiview}, where each node denotes one point set and each edge indicates the pair-wise registration of two connected nodes. Without the correspondence update, these approaches can diffuse the registration error over the graph of adjacent point sets. Despite their efficiency, these approaches only transfer registration errors among graph nodes and are unable to reduce the total registration errors.

Recently, Govindu and Pooja~\cite{govindu2004lie,govindu2014averaging} proposed the motion averaging algorithm for the multi-view registration. It can recover all rigid transformations simultaneously for multi-view registration from a set of relative motions, which can be estimated by the pair-wise registration. Given reliable and accurate relative motions \cite{li2014improved}, the motion averaging algorithm can achieve accurate multi-view registration. In practical applications, the relative motions obtained from the pair-wise registration should be confirmed \cite{govindu2006robustness,pankaj2016robust} so as to provide reliable and accurate relative motions to the motion averaging algorithm. Alternatively, the multi-view registration can also be cast into the problem of low-rank and sparse (LRS) matrix decomposition \cite{arrigoni2016global}. For some scan pairs with large overlap, the relative motions can be estimated and concatenated into a large matrix, which has missed data corresponding to the scan pairs with low overlapping percentages. By the LRS decomposition, the matrix can be completed and the multi-view registration results can be recovered. Compared to other approaches, it is more likely to be affected by the sparsity of the uncompleted matrix.

More recently, Evangelidis and Horaud \cite{evangelidis2017joint} held the assumption that all the points involved in multi-view registration are drawn from a single Gaussian mixture, thus formulating registration as a soft clustering problem. An expectation maximization (EM) algorithm is utilized to estimate both the mixture parameters and the rigid transformations that optimally align the point sets. Although this approach is accurate, lots of parameters need to be estimated and the computational cost is quite high.

In this paper, we propose a new multi-view registration algorithm formulated as a joint clustering and alignment problem. We build on well known clustering paradigm, i.e., the K-means algorithm. We first design the new objective function and then present the proposed multi-view registration algorithm. Given initially posed point sets, this algorithm starts with estimating the initial centroids. Then it performs the clustering operation, which consists in the assignment of each point to a single cluster and the update of all cluster centroids. As the multi-view point sets are not well aligned, the shape comprised by all the updated centroids can be used as a reference point set to sequentially estimate the rigid transformation for each point set by the pair-wise registration. To obtain the desired results, clustering and transformation estimation are alternately and iteratively applied to all point sets. For the 3D model reconstruction, the K-means clustering was initially introduced by Zhou \emph{et al}. in \cite{zhou2008accurate}, where the K-means clustering algorithm was utilized to detect and merge the corresponding points of the overlapping areas. It doesn't involve the update of rigid transformations and only can be viewed as a a post-processing step after multi-view registration step in 3D reconstruction.

%Although K-means algorithm can be viewed as a simplification of EM algorithm when all the covariances are fixed to identity and all the mixing coefficients are the same, the algorithm in \cite{Evangelidis17} doesn't work in this situation. Therefore, the proposed approach is different from the algorithm in \cite{Evangelidis17}.

The remaining of the paper is organized as follows. Section \ref{sec:kmeans_clustering} gives a brief introduction of K-means clustering algorithm. Following that, Section \ref{sec:kmeans_based_registration} proposes the novel approach for registration of multi-view point sets. In Section \ref{sec:experiments}, the proposed approach is tested on some benchmark data-sets. Section \ref{sec:conclusion} concludes this paper.

\section{K-means clustering}\label{sec:kmeans_clustering}

As an unsupervised learning method, K-means algorithm is effective and interpretable for the clustering. Given the number of clusters $K$ and the data set $X = \{ {\vec x_j}\} _{j = 1}^M$, the algorithm starts with initial estimation of the $K$ centroids $\{ \vec \mu _k^0\} _{k = 1}^K$, which can be either randomly selected or generated from the data sets. The clustering result is obtained from the iteration of the following two steps.

Step 1: data assignment
\begin{equation}
{c^q}(j) = \mathop {\arg \min }\limits_{k \in \{ 1,2,..,K\} } \left\| {{{\vec x}_j} - \vec \mu _k^{q - 1}} \right\|_2^2
\label{eq:LSn}
\end{equation}

Step 2: centroid update
\begin{equation}
\vec \mu _k^q = \frac{{\sum\nolimits_{j = 1}^M {\{ {c^q}(j) = k\} {{\vec x}_j}} }}{{\sum\nolimits_{j = 1}^M {\{ {c^q}(j) = k\} } }}
\end{equation}
Usually, good clustering results can be obtained by iteratively performing the above steps until some convergence criterion is achieved.

\section{K-means based multi-view registration}\label{sec:kmeans_based_registration}
%In this section, we present the multi-view registration problem. We first discuss the motivation of the proposed approach, and then proceed with the k-means based registration algorithm.

%\subsection{Motivation}
Given an accurate 3D scene model, the multi-view registration problem can be divided into multiple subproblems,
where each point set is pair-wise registered to the
accurate model, respectively. However, when scanning an unknown scene, there is no accurate model beforehand. What is available is just the coarse model reconstructed from the initially posed point sets, which contains most likely unregistered points as well as too many redundant points due to the overlapping areas among different point sets.

\begin{figure*}
\begin{center}
%\fbox{\rule{0pt}{2in} \rule{.9\linewidth}{0pt}}
\includegraphics[width= 0.8\linewidth]{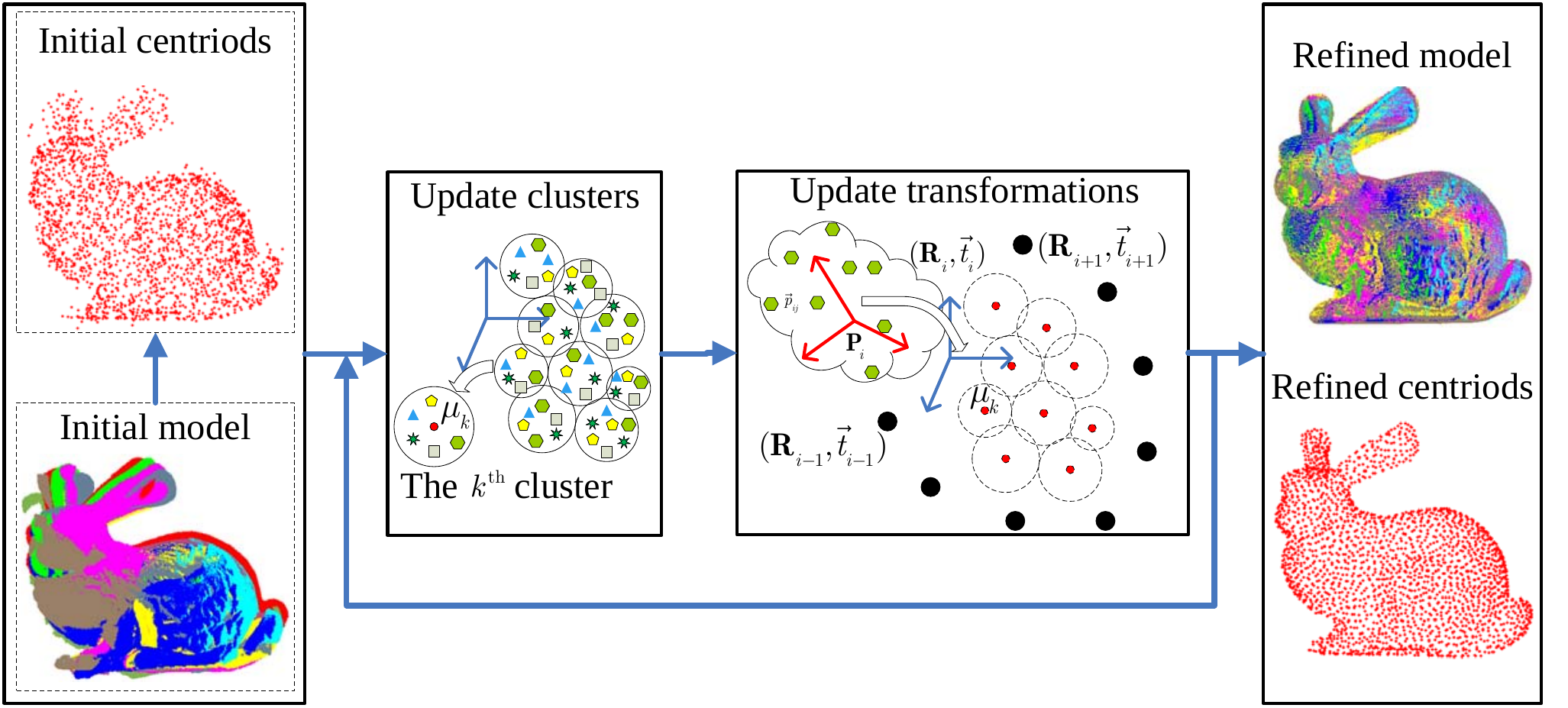}
\end{center}
    \caption{The proposed approach assumes that all points involved in multi-view registration are sampled from $K$ clusters. Given the multi-view point sets, the proposed approach starts with initial estimation for the $K$ cluster centroids. Then, clustering and transformation estimation are alternately and iteratively applied to all point sets until the stop criteria are met. The output contain both the multi-view registration and clustering results.}
\label{fig:Flowchart}
\end{figure*}

To deal with these two issues, we here suppose that all the points, once registered, are drawn from $K$ clusters, whose centroids can make up an accurate model. Therefore, if all the points can be well clustered, the accurate model can be constructed from the cluster centroids and the multi-view registration can be achieved by the pair-wise registration between the centroids and each point-set. Therefore, the multi-view registration can be viewed as the extended clustering problem, which incorporates a pair-wise registration solver. The flowchart of the proposed approach is displayed in Fig. \ref{fig:Flowchart}. Given the initial transformations that roughly align the point-sets, e.g. by any pair-wise registration scheme, the main steps of our approach are summarized as follows:

\begin{enumerate}
    \item Initialize centroids from initially aligned point sets, e.g. random selection of $K$ points.
    \item Assign each point to the nearest cluster and then update the centroids.
    \item Align each point-set against the centroids
    \item Iterate 2) and 3) until convergence
\end{enumerate}

\subsection{Problem formulation}
We are given $N$ point sets to register. Let  ${{\mathbf{P}}_i} = \{ {\vec p_{ij}}\} _{j = 1}^{{M_i}}({\vec p_{ij}} \in {\mathbb{R}^{3 \times 1}})$ denote the ${M_i}$ 3D points that belong to the $i$th point set and let $M = \sum\nolimits_{i = 1}^N {{M_i}}$ denote the total number of points. As mentioned, we assume that all the $M$ points, once registered, are drawn from $K$ clusters. Given initial rigid transformations $\{ {\mathbf{R}}_i^0,\vec t_i^0\} _{i = 1}^N$ that roughly align the point-sets (e.g. by pairwise registration), the goal of multi-view registration is to simultaneously group the $M$ points into $K$ clusters and estimate accurate rigid transformations $\{ {{\mathbf{R}}_i},{\vec t_i}\} _{i = 1}^N$ between each original point set and the centroids. This is an essential difference from \cite{evangelidis2017joint} which registers Gaussian means with virtual points sets produced by the original ones.

%Denote ${{\mathbf{P}}_i} = \{ {\vec p_{ij}}\} _{j = 1}^{{M_i}}({\vec p_{ij}} \in {\mathbb{R}^{3 \times 1}})$ as ${M_i}$ 3D points that belong to the $i$th point set and let $N$ be the number of point sets, where $M = \sum\nolimits_{i = 1}^N {{M_i}}$ denotes the total number of points. In the view of clustering, we can suppose all the $M$ points are drawn from $K$ clusters, which can be represented by the same number of centroids. Given initial rigid transformations $\{ {\mathbf{R}}_i^0,\vec t_i^0\} _{i = 1}^N$, the goal of multi-view registration is to simultaneously group these points into $K$ clusters and estimate accurate rigid transformations $\{ {{\mathbf{R}}_i},{\vec t_i}\} _{i = 1}^N$ between each point set and the reference frame. Without loss of generality, the reference frame can be attached to the first point set. Therefore, there is no need to estimate the first rigid transformation, which has always been fixed during the multi-view registration.

%\subsection{Multi-view registration based on the k-means clustering}
Given all the point sets $\{{\mathbf{P}}_i\}_{i = 1}^N$ and assuming that their points are samples drawn from $K$ clusters, represented via their centroids $\{ \vec \mu _k\} _{k = 1}^K$, the multi-view registration can be formulated as the following least-square (LS) problem:
\begin{equation}
\begin{gathered}
  \mathop {\arg \min }\limits_{{{\mathbf{R}}_i},{{\vec t}_i},c(ij) \in \{ 1,2,..,K\} } \sum\limits_{i = 2}^N {\sum\limits_{j = 1}^{{M_i}} {({w_{ij}} \cdot \left\| {{{\mathbf{R}}_i}{{\vec p}_{ij}} + {{\vec t}_i} - {{\vec \mu }_{c(ij)}}} \right\|_2^2)} }  \hfill \\
  {\text{s}}{\text{.t}}{\text{.    }}{{\mathbf{R}}_i}^T{{\mathbf{R}}_i} = {{\mathbf{I}}_3},\det ({{\mathbf{R}}_i}) = 1 \hfill \\
\end{gathered}
\label{eq:LS}
\end{equation}
where  ${\vec \mu _{c(ij)}}$ is the nearest cluster centroid of the transformed point $({{\mathbf{R}}_i}{\vec p_{ij}} + {\vec t_i})$ and ${w_{ij}}$ is a binary variable that quantifies the hard assignment. If the  $c(ij)$th cluster only contains the point that belongs to the  $i$th point set, we set ${w_{ij}} = 0$, otherwise ${w_{ij}} = 1$. Note that solving problem (\ref{eq:LS}) implies solving both the clustering problem and the registration problem, that is, the cluster centroids $\{ \vec \mu _k\} _{k = 1}^K$ are also unknown.

\subsection{Proposed multi-view registration algorithm}
K-means algorithm is very effective for the clustering. However, to solve problem (\ref{eq:LS}), it should be extended so as to cluster points and estimate the rigid transformations simultaneously.

Regardless of registration, initial centroids $\{ \vec \mu _k^0\} _{k = 1}^K$  are required for K-means. The initial centroids can be either randomly selected or generated from the data set. Given initial rigid transformations $\{ {\mathbf{R}}_i^0,\vec t_i^0\} _{i = 1}^N$, the coarse model can be reconstructed by all point sets involved in the multi-view registration as follows:
\begin{equation}
 {\mathbf{P}} \buildrel \Delta \over = \{ {\mathbf{R}}_i^0{\vec p_{ij}} + \vec t_i^0\} _{i = 1,j = 1}^{N,{M_i}}.
\end{equation}
Then, the initial centroids of all clusters $\{ \vec \mu _k^0\} _{k = 1}^K$ can be uniformly sampled from this coarse model. The shape comprised by all centroids is similar to the coarse model. The only difference is the resolution of points.

Being provided with initial rigid transformations $\{ {\mathbf{R}}_i^0,\vec t_i^0\} _{i = 1}^N$, the K-means clustering algorithm can be used to achieve the registration of multi-view point sets by iterations. In each iteration, the following three steps are included:

Step 1: Assign each point to a single cluster:
\begin{equation}
 {c^q}(ij) = \mathop {\arg \min }\limits_{k \in \{ 1,2,..,K\} } \left\| {{\mathbf{R}}_i^q{{\vec p}_{ij}} + \vec t_i^q - \vec \mu _k^q} \right\|_2^2.
 \label{eq:assign}
\end{equation}

Step 2: Update the centroids for all clusters:
\begin{equation}
 \vec \mu _k^q = \frac{{\sum\nolimits_{ij = 1}^M {\{ {c^q}(ij) = k\} ({\mathbf{R}}_i^q{{\vec p}_{ij}} + \vec t_i^q)} }}
{{\sum\nolimits_{ij = 1}^M {\{ {c^q}(ij) = k\} } }}.
\label{eq:update}
\end{equation}

Step 3: Sequentially estimate each rigid transformation:
\begin{equation}
\mathop {({\mathbf{R}}_i^q,\vec t_i^q)} = \arg \min \limits_{{{\mathbf{R}}_i},{{\vec t}_i}} \sum\limits_{j = 1}^{{M_i}} {w_{ij}^q\left\| {{{\mathbf{R}}_i}{{\vec p}_{ij}} + {{\vec t}_i} - \vec \mu _{{c^q}(ij)}^q} \right\|_2^2}.
\label{eq:calculate}
\end{equation}

As with the K-means algorithm, the above steps should be repeated until the iteration number $q$ exceeds the maximum value $Q$ or all the estimated rigid transformations have not drastically changed in successive iterations. Finally, it can simultaneously achieve the clustering and obtain the desired results for the registration of multi-view point sets.

In fact, Eq. (\ref{eq:assign}) is the nearest neighbor (NN) search problem, which can be efficiently solved by the search method based on k-d tree. After data assignment, each cluster centroid can be directly calculated from all the assigned points. Besides, Eq. (\ref{eq:calculate}) can be solved by the singular value decomposition (SVD) method \cite{zhu2016registration}. The main difficulty lies in the confirmation of each binary variable $w_{ij}^q$. In the multi-view registration, the $i$th point set covers some regions that other point sets cannot cover. If these regions are used to update the rigid transformation for itself in Eq. (\ref{eq:calculate}), the accuracy will go down. Therefore, these regions are invalid and should be eliminated. As these regions are only covered by one point set, the cluster lying in these regions would contain less points than other regions, which may be covered by more than two point sets. Hence, we can confirm the value of $w_{ij}^q$ according to the cardinality of the ${c^q}(ij)$th cluster. Here, if the cardinality of the ${c^q}(ij)$th cluster is less than four fifths of the mean value of all clusters, we set $w_{ij}^q = 0$ so as to eliminate the regions covered only by  $i$th point set. Otherwise, we set $w_{ij}^q = 1$.

%\subsection{Implementation}
Based on the above description, our multi-view registration algorithm is summarized in Algorithm \ref{alg:mvreg}. As seen, to efficiently register multiple point sets, all points in the coarse reconstructed model should be clustered so as to get the reasonable model comprised by the cluster centroids.
\begin{algorithm}[htbp]
        \caption{: Multi-view registration algorithm}
        %\begin{algorithmic}[1]
               \textbf{Input}: ${\Theta ^0} = \{ {\mathbf{R}}_i^0,\vec t_i^0\} _{i = 1}^N$, $K$ ,$\{ {{\mathbf{P}}_i}\} _{i = 1}^N$

            \quad Get the initial model by Eq. (4);

            \quad Sample the centroids $\{ \vec \mu _k^0\} _{k = 1}^K$ from the initial model;

            \quad $q = 0$;

       \quad \textbf{Repeat}

               \qquad $q = q + 1$;

               \qquad Do data assignment according to Eq. (\ref{eq:assign});

               \qquad Update all cluster centroids according to Eq. (\ref{eq:update});

               \qquad \textbf{for} i= 2:N

               \qquad \quad Calculate $({\mathbf{R}}_i^q,\vec t_i^q)$ according to Eq. (\ref{eq:calculate});

               \qquad \textbf{end}

        \quad \textbf{Until} ( $\Theta$'s change is negligible) or ($ q > Q$ )

                \textbf{Output}: $\Theta  = \{ {{\mathbf{R}}_i},{\vec t_i}\} _{i = 1}^N$
               \label{alg:mvreg}
    \end{algorithm}

The only parameter that affects the performance of the algorithm is the number of the clusters $K$, as discussed in Section \ref{sec:experiments}.

\subsection{Complexity}
This section discusses the complexity of the proposed
approach for registration of multi-view point sets.
Since our approach is proposed for registering multiple point sets, the total number $M$ of points is the central quantity.
For the multi-view registration, the maximum value of clustering iteration can be set as $Q$.
At each iteration, four steps are executed to estimate each transformation:

(1) Build the $k$-d tree. Before the estimation of all transformations, the $k$-d tree should be built once from all $K$ cluster centroids for the point assignment. This step results in a complexity of $O(K\log K)$.

(1) Assign each point to a single cluster. In this step, each point should be assigned to the nearest cluster centroid. To accelerate the assignment, the proposed approach utilizes the nearest neighbor search method based on the $k$-d tree, which leads to a complexity of $O(M\log K)$.

(2) Update the centroids for all clusters. By sequentially traversing each point, all the cluster centroids can be updated. Therefore, This step introduces a complexity of $O(M)$.

(3) Sequentially estimate each rigid transformation. As shown in Eq. (\ref{eq:calculate}), $M_i$ point pairs are used to estimate one rigid transformation, so the estimation of $N$ rigid transformations can introduce a complexity of $O(M)$.

Since $N$ transformations are required to be estimated, the complexity of our approach is shown in
Table \ref{tab:comp}.

\begin{table}  %表1
\footnotesize
\renewcommand\arraystretch{1}         %表格内部 2 倍行距离
\caption{Complexity of the different operations in each iteration of the proposed registration approach}
%\vspace{1.2 mm}                        %让标题与表格空出一个1.2毫米的行间距
\centering                            %居中显示
{\tabcolsep0.2in                     %列间距
\begin{tabular}{|c|c|}
 \hline
    Operation          & Complexity  \\
 \hline
    Building k-d tree  & $O(K\log K)$  \\
  \hline
Data assignment  &  $O(M\log K)$  \\
  \hline
Centroid update  &  $O(M)$  \\
  \hline
Transformation estimation & $O(M)$   \\
 \hline
\end{tabular}
}
\label{tab:comp}
\end{table}

\begin{table*}  %表1
\footnotesize
\renewcommand\arraystretch{1}         %表格内部 2 倍行距离
\caption{3D scans used for evaluation}
%\vspace{1.2 mm}                        %让标题与表格空出一个1.2毫米的行间距
\centering                            %居中显示
{\tabcolsep0.05in                     %列间距
\begin{tabular}{cccccccccc}
\toprule[1pt]
              & Bunny   & Armadillo & Buddha   & Dragon  & Angel    & Hand\\
  \hline
Number of views ($N$) &  10     & 12        &  15      &    15   &  36      &  36 \\
Total number of point ($M$)  & 1362272 & 307625    &  1099005 &  469193 & 2347854  & 1065575 \\
\bottomrule [1pt]
\end{tabular}
}
\label{tab:datasets}
\end{table*}

\section{Experiments}\label{sec:experiments}

To evaluate the performance of the proposed approach, a set of experiments were conducted on six data-sets from the Stanford 3D Scanning Repository \cite{levoy2005stanford}. Table \ref{tab:datasets} displays some information of these data-sets, including the number of views and total point number of each data-set. These data-sets contain the multi-view point sets and the ground truth of transformations for their registration. To reduce the runtime of multi-view registration, all the raw point sets have been down-sampled by a factor $S = 8$. For the comparison, the error of rotation matrix and translation is here defined as ${E_{\mathbf{R}}} = \tfrac{1}
{N}\sum\nolimits_{i = 1}^N {{{\left\| {{{\mathbf{R}}_{i,m}} - {{\mathbf{R}}_{i,g}}} \right\|}_F}} $ and ${E_{\vec t}} = \tfrac{1}
{N}\sum\nolimits_{i = 1}^N {{{\left\| {{{\vec t}_{i,m}} - {{\vec t}_{i,g}}} \right\|}_2}}$, where $({{\mathbf{R}}_{i,g}},{\vec t_{i,g}})$ denotes the ground truth of the $i$th rigid transformation and $({{\mathbf{R}}_{i,m}},{\vec t_{i,m}})$ indicates the one estimated by the multi-view registration approach.

During experiments, the maximum step for clustering was set as $Q = 500$. All the baselines adopted the nearest-neighbor search method based on $k$-d tree to establish point correspondences and were implemented in MATLAB. To produce unregistered point sets, we randomly perturb the ground trough transformations. Experiments were performed on a four-core 3.6 GHz computer with 8 GB of memory.

\subsection{Choice of $K$}
The number of cluster centers $K$ seems to affect the performance of multi-view registration. The actual number of the clusters is unknown and may vary with the data set to be registered.

\begin{figure}
\begin{center}
%\fbox{\rule{0pt}{2in} \rule{.9\linewidth}{0pt}}
\subfigure[]{\label{fig:KRotation}\includegraphics[width= 0.32\linewidth]{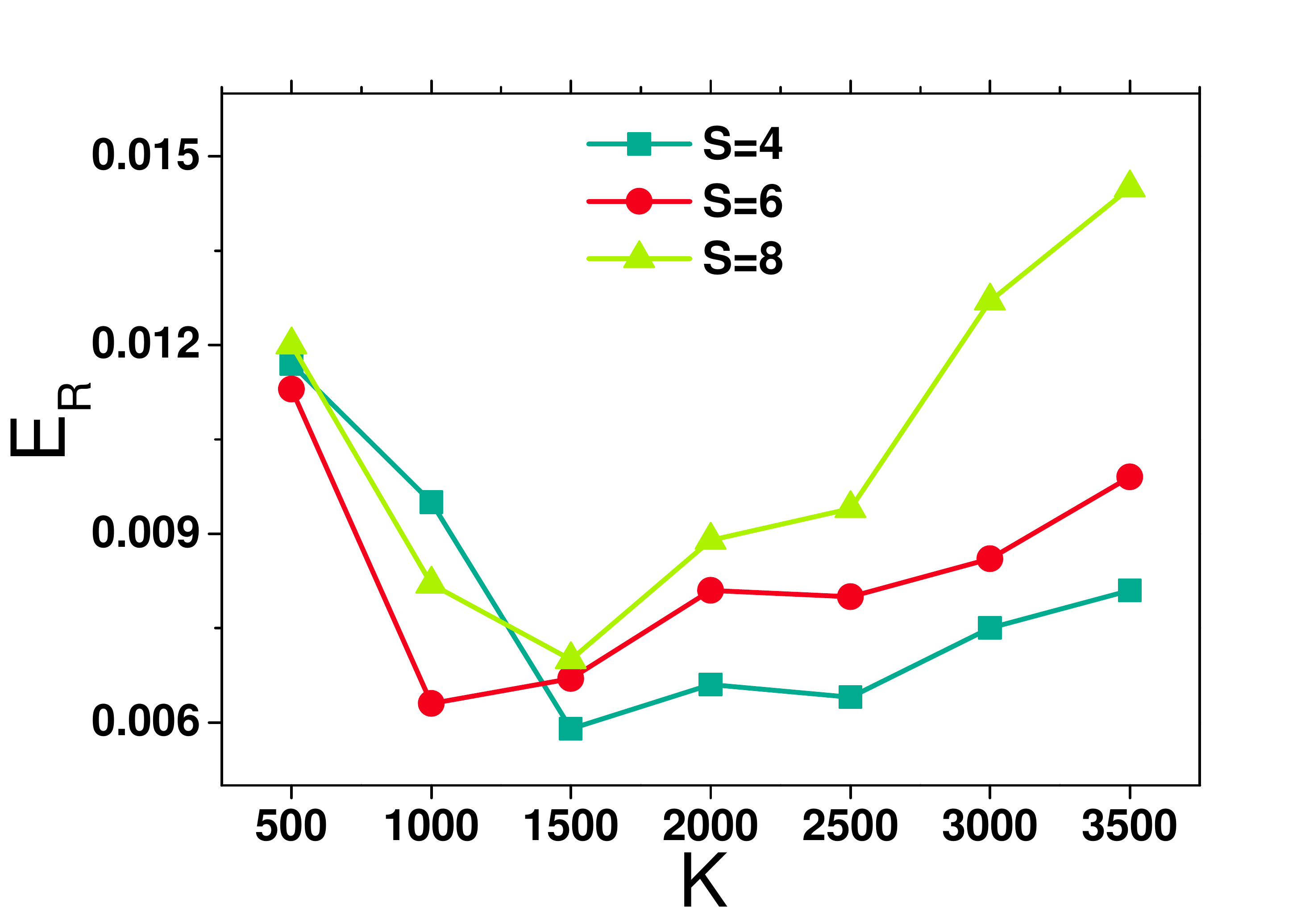}}
\subfigure[]{\label{fig:KTranslation}\includegraphics[width= 0.32\linewidth]{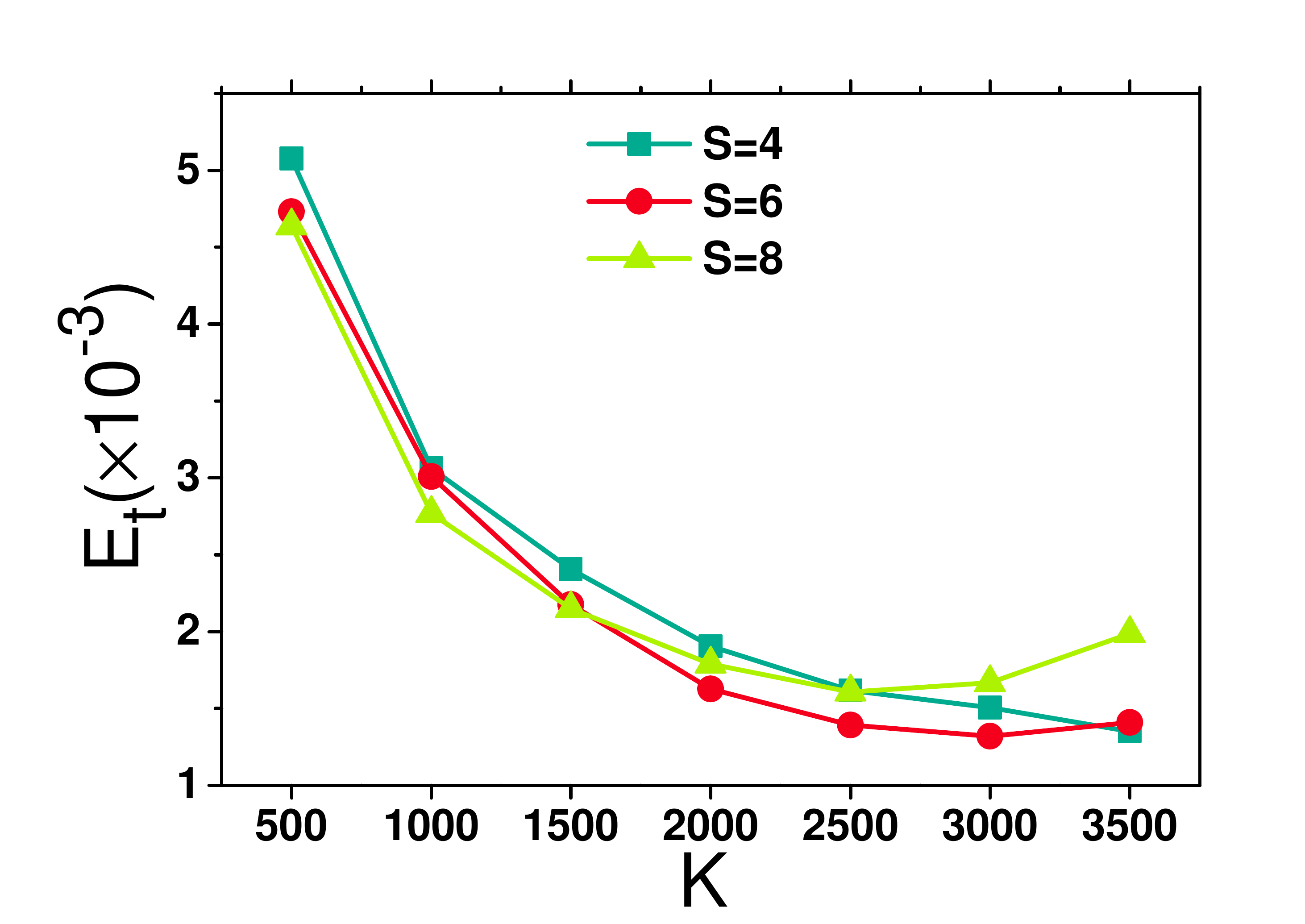}}
\subfigure[]{\label{fig:KTime}\includegraphics[width= 0.32\linewidth]{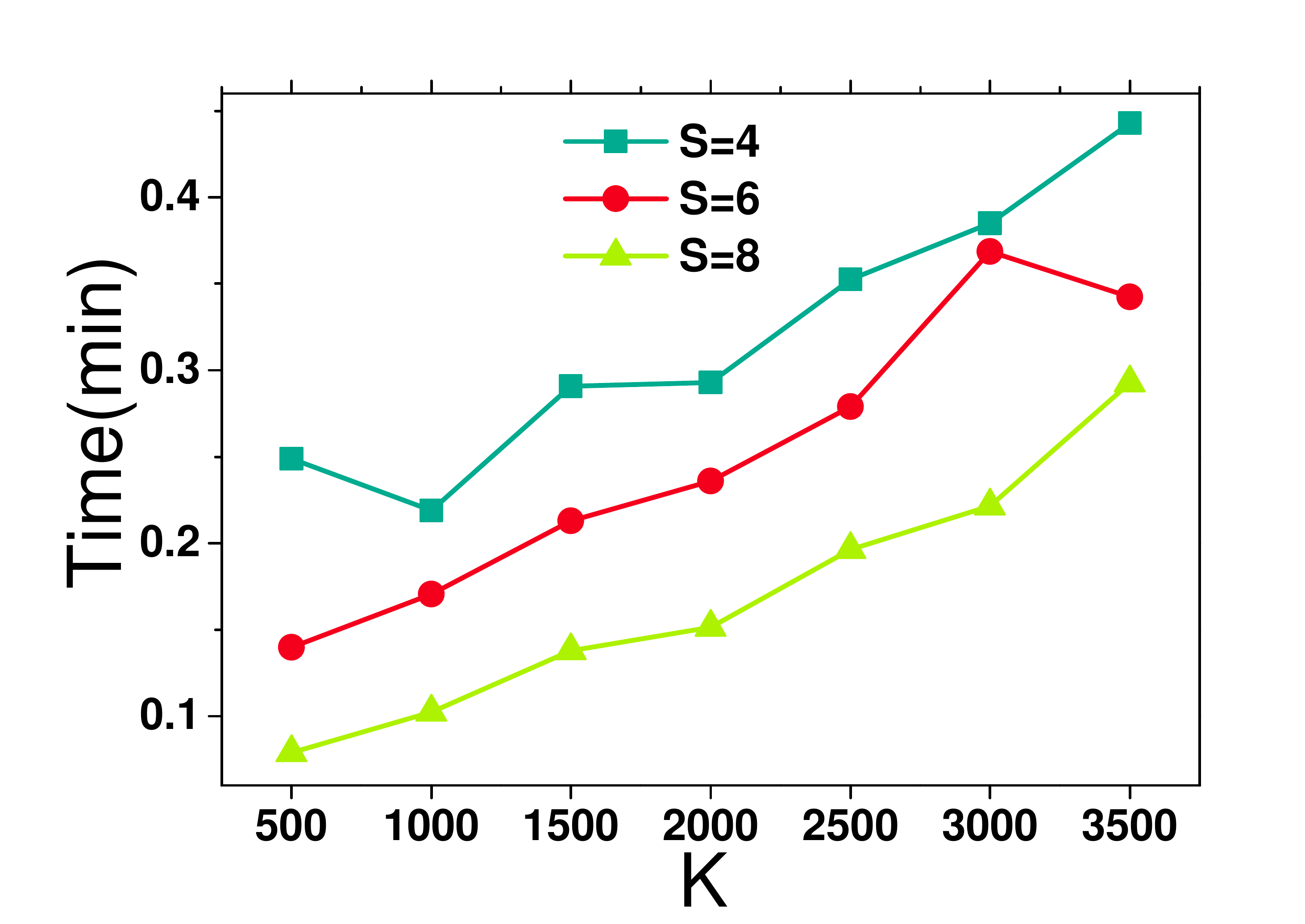}}
\end{center}
    \caption{ Mean error and run time of the proposed approach tested on Armadillo with different values of $K$. (a) Rotation error. (b) Translation error. (c) Runtime.}
\label{fig:SelectK}
\end{figure}

We conducted a group of experiments on Stanford Armadillo with different down-sampling factor $S$. To provide the initial registration parameters, a fixed perturbation was added to the ground-truth transformations of multi-view registration. Then, the proposed approach was tested with the number of cluster centers $K$ varying from $500$ to $3500$. To eliminate the randomness, $20$ Monte Carlo (MC) trials were conducted with respect to each value of $K$. Fig. \ref{fig:SelectK} shows comparison results of the proposed approach under different values of $K$.

As shown in Fig. \ref{fig:SelectK}, the minimum rotation error is obtained for the range $[1000,1500]$ of $K$ for all down-sampling factors. The translation error decreases with the increase of the number of cluster centers $K$, while it may increase again after $K=3500$. The run time obviously increases with $K$.
While a large value of $K$ might be better for accuracy reasons, a small value of $K$ leads to a more efficient registration. By the consideration of these two factors, we set $K=1500$ in the experiments below for the proposed registration algorithm.

\begin{figure*}
\begin{center}
%\fbox{\rule{0pt}{2in} \rule{.9\linewidth}{0pt}}
\includegraphics[width= 1.0\linewidth]{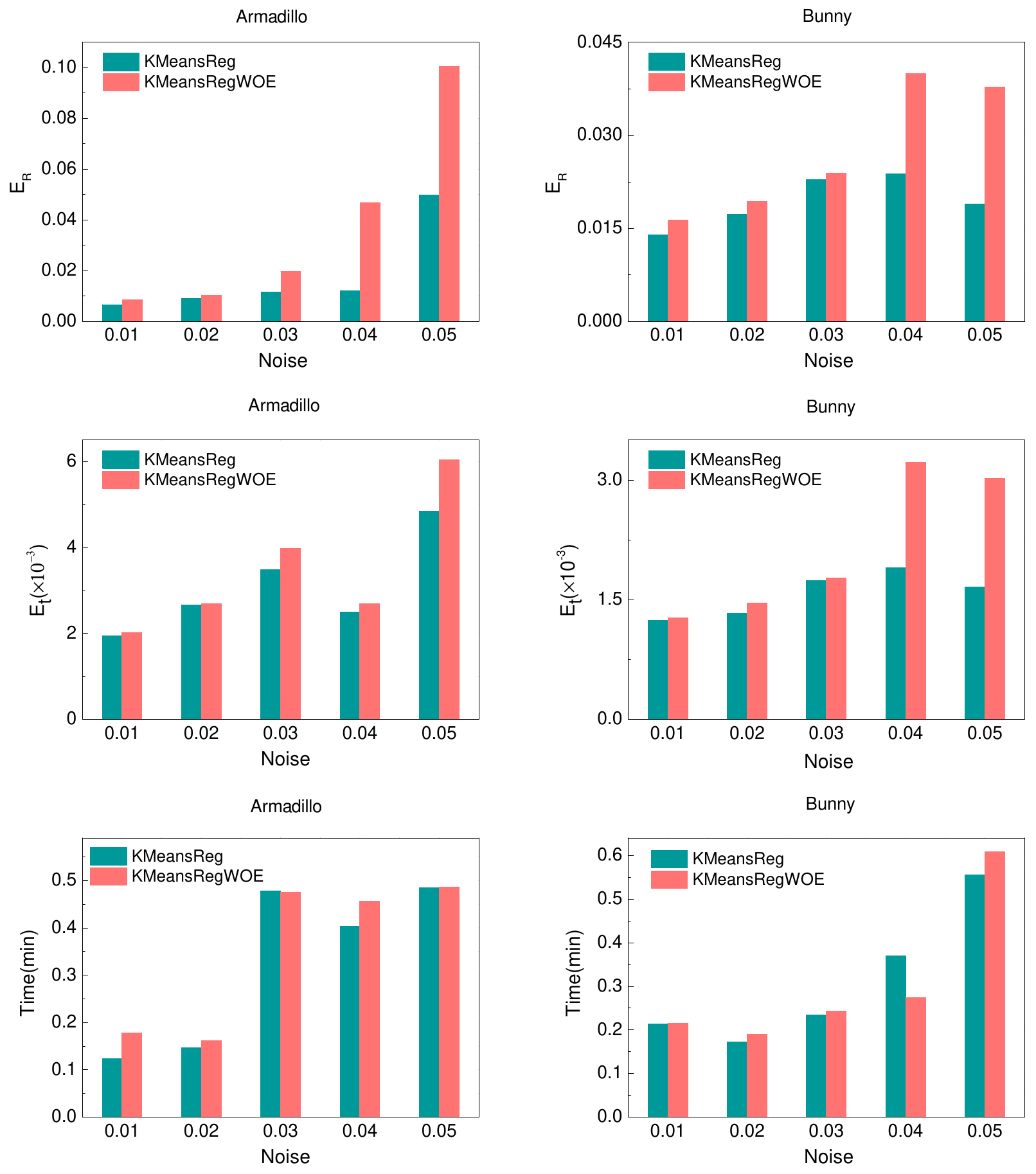}
\end{center}
    \caption{Comparison of different strategies for dealing with the invalid cluster centroids}
\label{fig:weight}
\end{figure*}

\subsection{Validation}
As mentioned before, the reconstructed model is comprised by the cluster centroids, which are initially sampled from all the aligned point sets. Since each point set covers some regions that other point sets cannot cover, the proposed approach suggests that these regions should be eliminated to estimate the rigid transformation of this point set itself. To verify the effectiveness of this strategy, we compare the proposed approach with and without elimination of the invalid cluster centroids, which are abbreviated as KmeansReg and KmeansRegWOE (Kmeans Registration Without Elimination), respectively. They were tested on Stanford Bunny and Armadillo under different perturbation levels, where five uniform distributed noises were added to ground-truth transformations, respectively. To eliminate the randomness, $10$ MC trials were conducted with respect to five perturbation levels. Fig. \ref{fig:weight} illustrates the mean of rotation error, translation error and the mean run time of the two different strategies.

\begin{table}  %表2
\footnotesize
\renewcommand\arraystretch{1}         %表格内部 2 倍行距离
\caption{Comparison of accuracy and runtime for different approaches}
%\vspace{1.2 mm}                        %让标题与表格空出一个1.2毫米的行间距
\centering                            %居中显示
{\tabcolsep0.01in                     %列间距
\begin{tabular}{cccccccc}
\toprule[1pt]
\multicolumn{1}{c}{Dataset} &                & Initial  &	CFTrICP \cite{zhu2014surface} &	MATrICP \cite{govindu2014averaging}  & LRS \cite{arrigoni2016global} & JRMPC \cite{evangelidis2017joint}   &	Ours \\
  \hline

\multirow{2}{*}{Bunny} & ${E_{\mathbf{R}}}$ & 0.0360 & \textbf{0.0032}	  & 0.0086	& 0.0082 & 0.0163  &	0.0111 \\
	
  	                     & ${E_{\vec t}(\times {10^{ - 3}})}$     & 2.5211 &	0.4181	  & \textbf{0.4176}	& 0.6115 & 1.9902  &	1.1439 \\
  	
                         & T(min)	                              & --      & 1.0828 &    0.3380	  & 0.9327	& 7.7482 & \textbf{0.2207}  \\
  \hline  	
\multirow{3}{*}{Armadillo} & ${E_{\mathbf{R}}}$ & 0.0375  & 0.0077 &	0.1088	  & 0.0650	& 0.0207 & \textbf{0.0049} \\
	
                         & ${E_{\vec t}(\times {10^{ - 3}})}$     & 4.5276 &	\textbf{0.7717}	  & 7.8606	& 7.4092 & 0.8416  &	2.1517 \\

	                     & T(min)	          & --  &    0.7941	  & 3.8178	& $ > 10$ & 6.9728 &   \textbf{0.1426} \\
   \hline
\multirow{3}{*}{ Buddha} & ${E_{\mathbf{R}}}$ & 0.0334 & 	0.0333	  & \textbf{0.0107}	& 0.0310	 & 0.0213   & 	0.1320 \\
	
       	                 & ${E_{\vec t}(\times {10^{ - 3}})}$     & 2.7316 &	2.1410	  & 1.0994	& 1.4452 & \textbf{0.7771}	&    1.8696 \\

	                     & T(min)             &	-- &	$ > 10$	  & 2.0492	& 4.8167 & $ > 10$	&   \textbf{0.7671} \\
  \hline
\multirow{3}{*}{ Dragon }     & ${E_{\mathbf{R}}}$ & 0.0439 &	0.0429	  & 0.2770	& 0.2539 & 0.0164	&    \textbf{0.0106} \\
	
        	             & ${E_{\vec t}(\times {10^{ - 3}})}$     & 5.6432 &	3.8421	  & 39.6289	& 36.3570 & \textbf{1.2654}   &	2.5963 \\

	                     & T(min)	          & -- &	7.6386	  & 3.2058	& $ > 10$ & $ > 10$	&   \textbf{0.3042} \\
  \hline
\multirow{3}{*}{Angel}      & ${E_{\mathbf{R}}}$ & 0.0382 &	0.0305	  & \textbf{0.0022}	& 0.0072 & 0.0105	&    0.0045 \\
	
        	             & ${E_{\vec t}(\times {10^{ - 3}})}$     & 184.7460 &	410.3960	  & \textbf{51.8375}	& 72.2118 & 66.2194  &	94.2439 \\

	                     & T(min)	          & -- &	$ > 10$	  & 7.7475  	& $ > 10$ & $ > 10$	&   \textbf{2.2303} \\
  \hline	
\multirow{3}{*}{Hand}    & ${E_{\mathbf{R}}}$ & 0.0423 &	0.0518   & 0.0712	& 0.0569 &	0.0175  &	\textbf{0.0097} \\
	
                         & ${E_{\vec t}(\times {10^{ - 3}})}$     & 227.1540 &	153.5512	  & 261.3473  & 206.0000 &	96.6633  &	\textbf{85.8901} \\

                         & T(min)             &	-- &	$ > 10$	  & $ > 10$	& $ > 10$ &	$ > 10$	&   \textbf{1.3036} \\
\bottomrule [1pt]
\end{tabular}
}
\label{tab:com}
\end{table}

As shown in the Fig. \ref{fig:weight}, the proposed approach without elimination of the invalid cluster centroids can always obtain the less accurate registration results under varied noise levels for different data-sets. This is because the pair-wise registration of two identical point sets can only obtain the identity matrix. Since the $i$th rigid transformation should be estimated by aligning the  $i$th point set to the model comprised by all the cluster centroids, these cluster centroids located in the regions only covered by the $i$th point set should be eliminated. Otherwise, the registration accuracy drops. Therefore, the proposed approach with elimination of the invalid cluster centroids can obtain more accurate registration results. Besides, the execution time of these two strategies are approximately equal. As a result, it is reasonable to discard the invalid cluster centroids so as to accurately estimate the transformations.

\begin{figure}
\begin{center}
%\fbox{\rule{0pt}{2in} \rule{.9\linewidth}{0pt}}
\includegraphics[width= 1.0\linewidth]{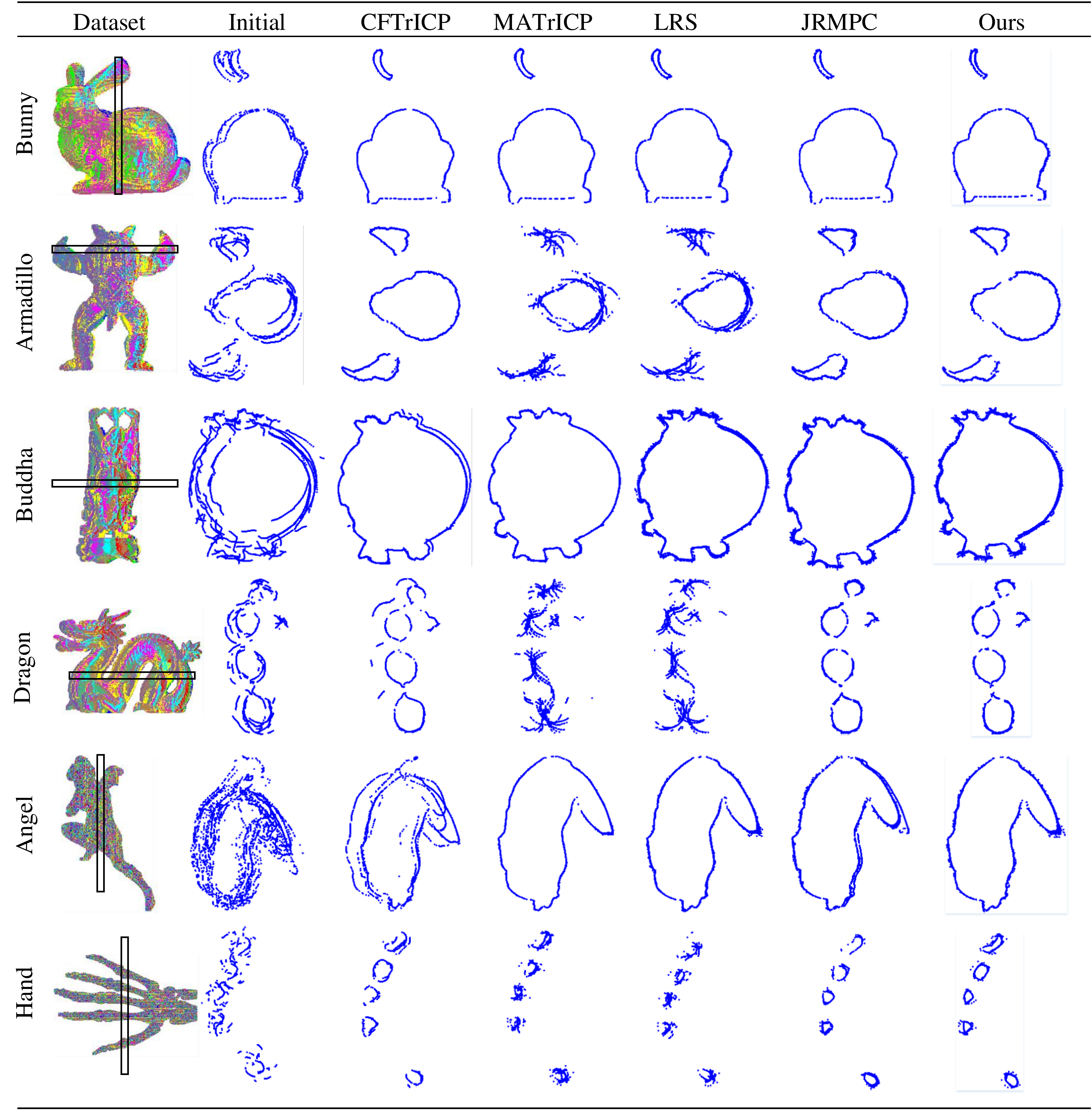}
\end{center}
    \caption{Cross-section of multi-view registration results}
\label{fig:crosssection}
\end{figure}

\subsection{Comparison}
We here compare the proposed algorithm against four state-of-the-art approaches: the coarse-to-fine TrICP approach \cite{zhu2014surface}, the motion averaging TrICP approach \cite{govindu2014averaging}, the approach based on the low-rank and sparse decomposition \cite{arrigoni2016global} as well as the joint registration approach \cite{evangelidis2017joint}, which are abbreviated as CFTrICP, MATrICP, LRS, and JRMPC, respectively. Results are reported in run time, the error of rotation matrix and translation.

\subsubsection{Accuracy and efficiency}
As the locally convergent registration approaches,all these five approaches require the initial registration parameters. As mentioned, initial registration parameters were generated by perturbing the ground truth of rigid transformations with some noise. Subsequently, all the competed approaches start from the same initial parameters for the registration of multi-view point sets. For comparison, Table \ref{tab:com} shows the registration error and runtime of all competed approaches for different data sets, where the bold number denotes the best performance.To evaluate the registration accuracy in a more intuitive way, Fig. \ref{fig:crosssection} displays the multi-view registration results in the form of cross-section.

As shown in Table \ref{tab:com}, the proposed approach is the most efficient one among these competed approaches. Although JRMPC is also a clustering based registration approach, it is more time-consuming than the proposed approach owing to the soft assignment strategy and to the larger number of parameters that need estimating. Note that, the proposed approach only requires to establish correspondences between each point set and the model comprised by the cluster centroids. More specifically, it only requires to assign a single cluster centroid to one point in each point set. While, the CFTrICP should establish correspondences between each point set and the model constructed by other aligned point sets. Besides, both MA-TrICP and LRS should establish correspondences between point set pairs, which contains certain percentage of overlaps. Since the number of the cluster centroids are much less than the number of point involved in multi-view registration, the establishment of correspondences, which is the most time-consuming operation in the multi-view registration, is much faster in the proposed approach.

As shown in Table \ref{tab:com} and Fig. \ref{fig:crosssection}, the proposed approach is not very sensitive to noises and it can always obtain acceptable accuracy compared to other baselines. Since the proposed approach utilizes the model comprised by all cluster centroids for the estimation of each transformation, the resolution of this model is much lower than the model constructed by raw point sets. Therefore, the proposed approach may not always obtain the most accurate registration results. But other competed approaches may fail to achieve the multi-view registration without good registration parameters. Therefore, the proposed approach is comparable to other state-of-the-art approaches in terms of accuracy.

\subsubsection{Robustness}

\begin{table}  %表3
\footnotesize
\renewcommand\arraystretch{1}         %表格内部 2 倍行距离
\caption{Robustness comparison under varied noise levels}
\vspace{1.2 mm}                        %让标题与表格空出一个1.2毫米的行间距
\centering                            %居中显示
{\tabcolsep0.02in                     %列间距
\begin{tabular}{cccccccccc}
\toprule[1pt]
\multicolumn{1}{c}{Terms}  &        & CFTrICP \cite{zhu2014surface}  &	MATrICP \cite{govindu2014averaging} &	LRS \cite{arrigoni2016global}  & JRMPC \cite{evangelidis2017joint} &  Ours\\
  \hline
  	
\multirow{3}{*}{[-0.01,0.01]} & ${E_{\mathbf{R}}}$ & 0.0106 &	0.0745	  & 0.0507	& 0.0093 & \textbf{0.0061} \\
	
                         & ${E_{\vec t}(\times {10^{ - 3}})}$     & \textbf{0.8539} &	4.9418	  & 6.7188	& 2.1287 & 2.0236\\

	                     & T(min)	          & 1.0482 &   1.5948 	  & 4.6892	& 6.9014 & \textbf{0.1612} \\
 \hline
\multirow{3}{*}{[-0.02,0.02]} & ${E_{\mathbf{R}}}$ & 0.0428 &   0.1157	  & 0.0566  & 0.0156 &  \textbf{0.0092}\\
	
       	                 & ${E_{\vec t}(\times {10^{ - 3}})}$      & 2.3353 &   7.5286	  & 7.0797	& 2.5197 & \textbf{2.3219} \\

	                     & T(min)             &	 1.3056 &	1.6660	  & 7.9593	& 6.9335  & \textbf{0.1717} \\
  \hline
\multirow{3}{*}{[-0.03,0.03]}     & ${E_{\mathbf{R}}}$ & 0.0676 & 0.1450      &  0.0627	&  0.0259 & \textbf{0.0159} \\
	
        	             & ${E_{\vec t}(\times {10^{ - 3}})}$     & 2.8578 & 8.6094	  & 7.1143	&  3.1651  & \textbf{2.6353} \\

	                     & T(min)	          & 1.7885 & 1.7847	  & 8.6718	&  6.9945  &  \textbf{0.2475} \\
 \hline
\multirow{3}{*}{[-0.04,0.04]}     & ${E_{\mathbf{R}}}$ & 0.0845 & 0.1523     &  0.0636 &  0.0664  &  \textbf{0.0438}\\
	
        	             & ${E_{\vec t}(\times {10^{ - 3}})}$      & 4.5926 & 8.6997	  & 7.3976	&  6.0247 & \textbf{3.2839}\\

	                     & T(min)	          & 1.7990 & 1.8480	  & 10.1937	&  7.0830 & \textbf{0.3005} \\

\bottomrule [1pt]
\end{tabular}
}
\label{tab:rub}
\end{table}

\begin{figure*}
\begin{center}
%\fbox{\rule{0pt}{2in} \rule{.9\linewidth}{0pt}}
\subfigure[]{\label{fig:RobRotation}\includegraphics[width= 0.45\linewidth]{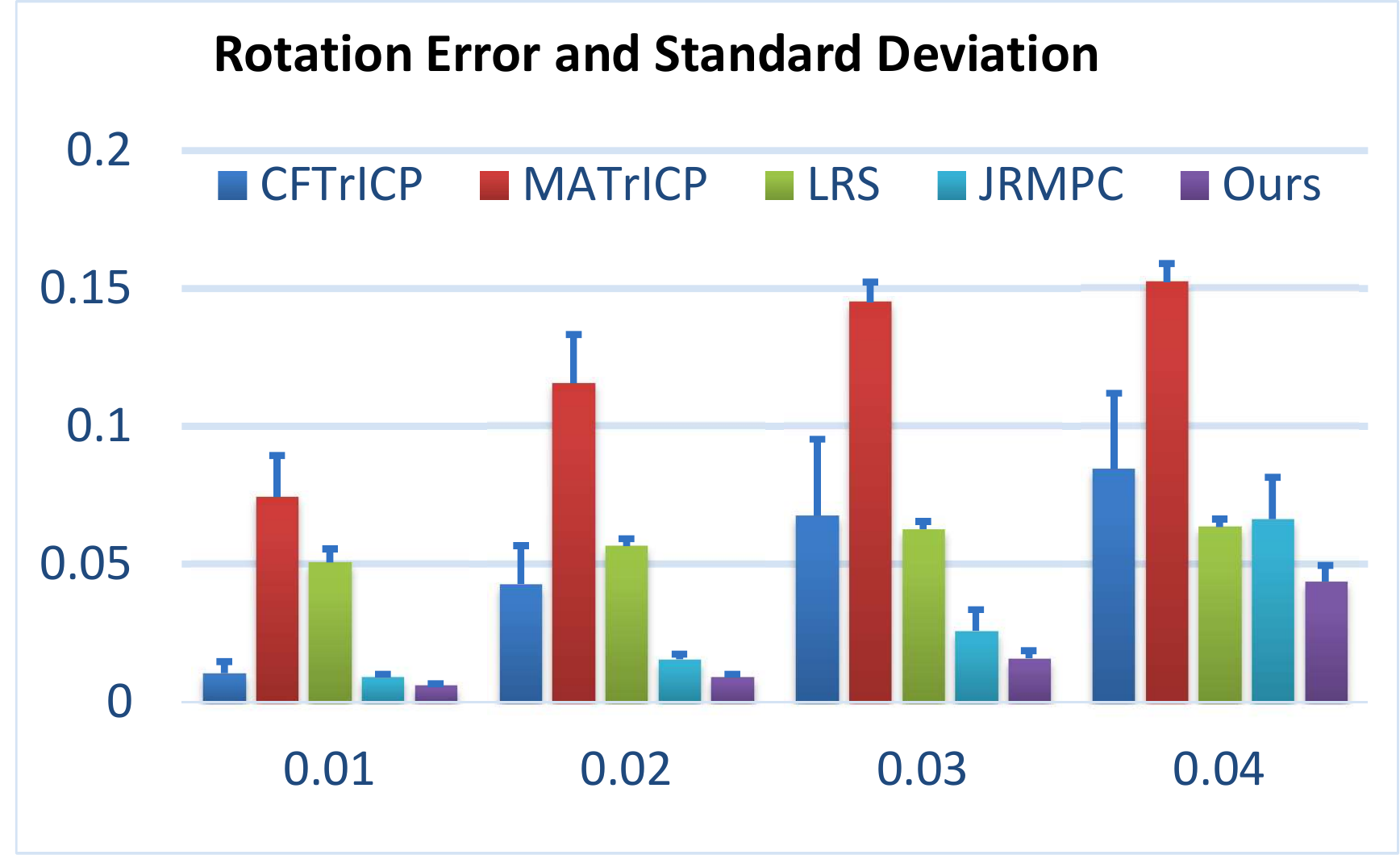}}
\subfigure[]{\label{fig:RobTranslation}\includegraphics[width= 0.45\linewidth]{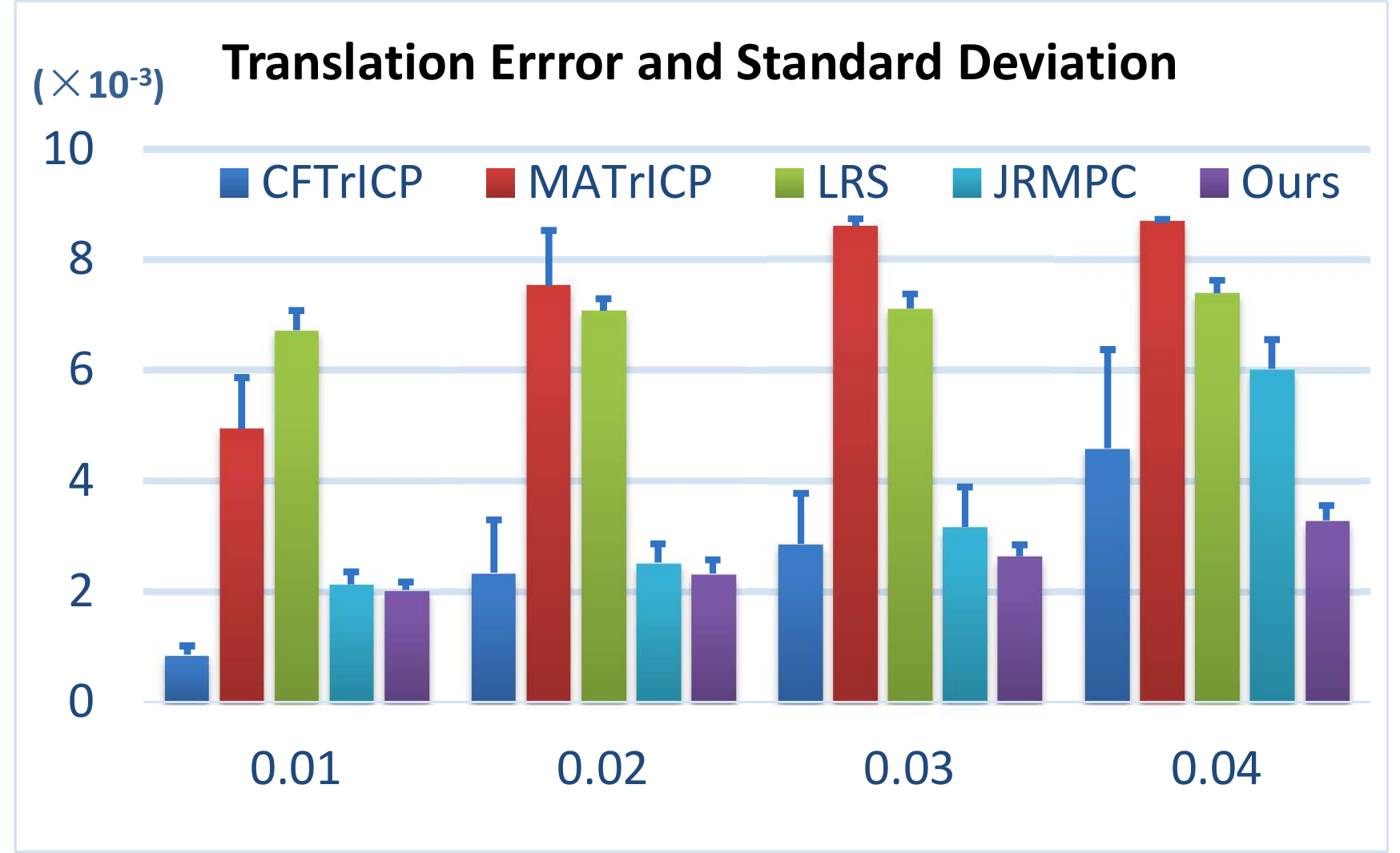}}
\end{center}
    \caption{Mean error and standard Deviation with respect to different noise levels}
\label{fig:rob_error}
\end{figure*}

To further verify its robustness, the proposed approach was tested on Stanford Armadillo under different transformation noise levels, where values sampled from five uniform distributions were added to ground-truth transformations, respectively. To eliminate the randomness, $10$ MC trials were conducted with respect to five noise distributions for all competed approaches. Table \ref{tab:rub} shows the average of rotation error, the average of translation error, and the average run time for these approaches. To compare the robustness in a more intuitive way, Fig. \ref{fig:rob_error} displays the mean error and the corresponding standard deviation for these approaches.

%\begin{figure}
%\begin{center}
%%\fbox{\rule{0pt}{2in} \rule{.9\linewidth}{0pt}}
%\includegraphics[width= 0.85\linewidth]{Rob_time.pdf}
%\end{center}
%    \caption{Runtime with respect to different noise levels}
%\label{fig:rob_time}
%\end{figure}

As shown in Table \ref{tab:rub} and Fig. \ref{fig:rob_error}, the proposed approach can obtain the most robust registration results under varied initial registration. With the increase of noise level, the registration errors of the proposed approach increase gradually and its standard deviation is always small. Although the CFTrICP and JRMPC perform well under low noise levels, their registration error raises at higher noise levels. Besides, the standard deviation of the CFTrICP increases sharply with the noise levels. Meanwhile, both MATrICP and LRS approaches have difficulties to obtain accurate registration for the Stanford Armadillo under all noise levels. Therefore, the proposed approach seems to be the most robust one among all the competed approaches. Besides, its run time is far less than other three competed approaches.

\section{Conclusion}\label{sec:conclusion}
This paper proposes a novel approach for registration of multi-view point sets. To the best of our knowledge, it is the first time applying the K-means clustering to the multi-view registration problem. The proposed approach has been tested on the Stanford 3D Scanning Repository and
the experimental results demonstrate that it can achieve the multi-view registration of initially posed point sets with good
efficiency and robustness. Similar to most of the related approaches, the proposed approach requires initial parameters for the multi-view registration.

Our future work will investigate how to estimate the initial rigid transformations so as to automatically achieve the multi-view registration of unordered point sets without any prior information.

%\appendices

\section*{Acknowledgment}
This work is supported by the National Natural Science Foundation of China under
Grant nos. 61573273, 61573280 and 61503300. We would like to thank Federica Arrigoni for providing the MATLAB implementation of~\cite{arrigoni2016global}.


\section*{Reference}

\bibliographystyle{elsarticle-num}
\bibliography{mybibfile}


%\begin{thebibliography}{00}


%\end{thebibliography}
\end{document}